\documentclass[runningheads]{llncs}
\usepackage{graphicx}
\usepackage[dvipsnames]{xcolor}
\usepackage{hyperref}
\usepackage{enumitem}
\usepackage{bm}
\usepackage{booktabs}
\usepackage{multirow}
\usepackage{adjustbox}
\usepackage{comment}
\usepackage{subcaption}
\usepackage[inkscape=true,inkscapeformat=png]{svg}
\newcommand\dataset[0]{$\mathcal{D}$}
\newcommand\datasetbalanced[0]{$\mathcal{D}^{bal}$}
\newcommand\datasetfewshot[1]{$\mathcal{F}^{#1}$}

\begin{document}
\title{Soft-prompt tuning to predict lung cancer using primary care free-text Dutch medical notes}
%
%\titlerunning{Abbreviated paper title}
% If the paper title is too long for the running head, you can set
% an abbreviated paper title here
%
\author{Auke Elfrink\inst{1,2}
\orcidID{0009-0006-1805-5597}
\and
Iacopo Vagliano\inst{2}
\orcidID{0000-0002-3066-9464}
\and
Ameen Abu-Hanna\inst{2}
\orcidID{0000-0003-4324-7954}
\and \\
Iacer Calixto\inst{2}
\orcidID{0000-0001-6244-7906}
}
\authorrunning{A. Elfrink et al.}
% First names are abbreviated in the running head.
% If there are more than two authors, 'et al.' is used.
%
\institute{
Informatics Institute, University of Amsterdam, Amsterdam, Netherlands. \and
Amsterdam UMC, University of Amsterdam, dept Medical informatics, Amsterdam Public Health research institute, Meibergdreef 9, Amsterdam, Netherlands 
\email{auke.elfrink@gmail.com, \{i.vagliano,a.abu-hanna\}@amsterdamumc.nl}
}
\maketitle              % typeset the header of the contribution
\begin{abstract}
We investigate different natural language processing (NLP) approaches based on contextualised word representations for the problem of early prediction of lung cancer using free-text patient medical notes of Dutch primary care physicians.
Because lung cancer has a low prevalence in primary care, we also address the problem of classification under highly imbalanced classes.
Specifically, we use large Transformer-based pretrained language models (PLMs) and investigate: 
1) how \textit{soft prompt-tuning}---an NLP technique used to adapt PLMs using small amounts of training data---compares to standard model fine-tuning;
2) whether simpler static word embedding models (WEMs) can be more robust compared to PLMs in highly imbalanced settings; and
3) how models fare when trained on notes from a small number of patients.
We find that
1) soft-prompt tuning is an efficient alternative to standard model fine-tuning;
2) PLMs show better discrimination but worse calibration compared to simpler static word embedding models as the classification problem becomes more imbalanced; and
3) results when training models on small number of patients are mixed and show no clear differences between PLMs and WEMs.
All our code is available open source in \url{https://bitbucket.org/aumc-kik/prompt_tuning_cancer_prediction/}.\footnote{A short version of this paper has been published at the 21st International Conference on Artificial Intelligence in Medicine (AIME 2023).}

\end{abstract}
\section{Introduction}
\label{sec:ïntro}
In the Netherlands, general practitioners (GPs) act as a gatekeeper between patients and the secondary healthcare system: GPs decide who should be referred to specialist care.
They also oversee the same patients for many years.
A patient's visit to their GP is registered in narrative form as \textit{free-text notes}, and collectively these provide a unique lens into years of a patient's health and medical history.
In previous work, Luik et al. \cite{torec-et-al2021w2vlungcancer} have used free-text patient notes in the early prediction of lung cancer with \textit{context-independent} word embedding methods (WEMs) and a simple logistic regression objective, with promising preliminary results.
Ideally, one would like to use state-of-the-art contextualised pretrained language models (PLMs) instead of simpler context-independent WEMs, since PLMs typically show improved performance across a range of different Natural Language Processing tasks~\cite{wang-etal-2018-glue,NEURIPS2019_superglue} including tasks in the (bio)medical domain \cite{fries2022bigbio,Naseem2022}.
However, there are at least two important reasons that make using PLMs difficult for predicting lung cancer from patient free-text notes:
i) Even though PLMs are language models pretrained on large amounts of raw text using self-supervised learning~\cite{delobelle-etal-2020-robbert,Verkijk_Vossen_2021}, such models are data-hungry and typically need large amounts of supervised training data to achieve good predictive performance on any specific task.
Conversely, lung cancer has a low prevalence (about $0.4\%$ in our dataset), which makes adapting PLMs to this task difficult.
ii) A single patient may have many years worth of health records, meaning possibly thousands of words in their medical notes.
A PLM's complexity is quadratic with respect to input length, making them non-trivial to apply in this use-case.

In this work, we propose methods to deal with the aforementioned issues by addressing the following research questions.
\textbf{RQ1.} \textit{How does soft-prompt tuning compare to standard model fine-tuning in terms of discrimination and calibration?}
In this work, we use \textit{soft prompt-tuning}~\cite{lester-etal-2021-power,liu-etal-2022-p}, an NLP technique used to adapt PLMs using small amounts of training data, and compare it to standard model fine-tuning.
Overall, we find that soft-prompt tuning compares favourably to standard model fine-tuning, but similar to fune-tuning when models are trained on small numbers of patients.
\textbf{RQ2.} \textit{How does the number of patients used for model training affect model performance in terms of discrimination and calibration?}
We empirically evaluate models trained on very small numbers of patients---a setting also known as \textit{few-shot learning}~\cite{wang2020fewshot}---and models trained on datasets with different degrees of class imbalance.
We report discrimination and calibration metrics, since both should be assessed for clinical prediction models~\cite{PMID:30596876}.
\textbf{RQ3.} \textit{How do PLMs compare to simpler static WEMs in terms of discrimination and calibration?}
Overall, contextualised PLMs outperform static WEMs in terms of discrimination, but WEMs show better calibration in both balanced and imbalanced classification settings.
However, results we obtain for few-shot experiments show mixed results, indicating that WEMs may be an alternative to PLMs in this setting.

%The remainder of this paper is as follows.
%In Section~\ref{sec:methods} we introduce the methods we use to answer the research questions RQ1--2 and we provide details on the patients we use in our experiments, who are patients from the primary care network associated with the Amsterdam University Medical Centers in Amsterdam, the Netherlands.
%In Section~\ref{sec:results} we report our main findings and discuss the most important results.
%In Section~\ref{sec:related} we contextualise our work within the existing research in pretrained language models in the (bio)medical domain.
%Finally, in Section~\ref{sec:conclusions}, we present our main conclusions and provide some avenues for future work.
\section{Methodology}
\label{sec:methods}

We now provide details on the patient selection and data cleaning before elaborating on our experimental settings.

\subsection{Data}
\label{sec:data}

We have unique access to data from patients of General Practitioners (GPs) that are associated with the two hospitals of the Amsterdam University Medical Centers, the Free University Medical Center (VUMC) and the Academic Medical Center of the University of Amsterdam (AMC).
Our patient cohort has an entry date of 01/01/2002 and an exit date of 31/12/2020, and includes patient demographics and free-text GP notes for every patient visit to their GP.
Henceforth, we only consider free-text notes that fall within the entry and exit dates.
We include patients who conform with the following criteria:
\textit{i)} the patient must be at least 40 years-old---i.e., we define the patient age as the date of their most recent free-text note minus their year of birth---,
\textit{ii)} the patient must have at least one valid free-text note.\footnote{All notes were anonymized with a modified version of the DEDUCE algorithm \cite{menger_deduce_2018}.}
Furthermore, we define how old a patient's free-text note is by the number of days between the date of the note and that patient’s last available note.

A free-text note is considered valid depending on whether the note belongs to a patient diagnosed with lung cancer or not.
For a free-text note of a patient diagnosed with lung cancer to be considered valid, the note must be dated at least 150 days ($\sim 5$ months) and at most 730 days ($\sim 2$ years) before the date of the diagnosis.
In case the patient is not diagnosed with lung cancer, the note must be dated at most 730 days ($\sim 2$ years) after the date of the patient's last available note.
Finally, we also exclude any notes that are undated, or have a date outside of the collection period.\footnote{For example, a reminder for a 5-year checkup.} 
We use these cut-off periods to ensure the predictive value of the model, and since GPs typically start to suspect lung cancer from about 4 months before diagnosis, and symptoms are expected to be present from 4 months to 2 years before diagnosis \cite{iyen_using_2013}.
Finally, we denote by \dataset{} the set of patients after we apply all the above mentioned procedures.

We establish lung cancer diagnoses for the patients in our cohort by linking to a central database maintained by \textit{Integraal Kankercentrum Nederland} (Integrated Cancer Center, IKNL).
Henceforth, we refer to patients with a lung cancer diagnosis as `positive', and all other patients as `negative'.

\subsubsection{Data splits}
\label{sec:data_splits}

In Table~\ref{tab:data_statistics} we show key characteristics for the different datasets used in our experiments: \datasetbalanced, $\mathcal{D}^{1:10}$, $\mathcal{D}^{1:100}$, $\mathcal{D}^{1:250}$, \datasetfewshot{2}, \datasetfewshot{4}, \datasetfewshot{8}, \datasetfewshot{16}, \datasetfewshot{32}, \datasetfewshot{64}, and \datasetfewshot{128}.
Train, validation, and test splits are always \textit{stratified}, i.e., they inherit the same ratio of patients with/without cancer as the original dataset.
Below we describe different data splits we use in different experiments in more detail.

\paragraph{Balanced dataset}
We first create the subset \mbox{\datasetbalanced{} $\subset$ \dataset{}} with the same number of positive and negative patients.
In \datasetbalanced{} we include all $1,733$ positive patients in \dataset{} and an equal number of negative patients, i.e., we randomly subsample patients without lung cancer from \dataset{} to have a balanced dataset.

\paragraph{Imbalanced test sets}
To study the effect of class imbalance in our models, we build test sets where the class imbalance between the positive and negative class (which in \datasetbalanced{} has a $1$:$1$ ratio) becomes closer and closer to the true ratio in the dataset population (which has approximately a $1$:$250$ ratio).
More concretely, we propose different \textit{test sets} $\mathcal{D}^{pos:neg}$ with different positive-to-negative ratios, and build the test sets $\mathcal{D}^{1:10}$, $\mathcal{D}^{1:100}$, and $\mathcal{D}^{1:250}$.

\paragraph{Few-shot datasets}
Finally, we also build few-shot datasets \datasetfewshot{k} $\subset$ \datasetbalanced{} that include a small number of patients, and where $k$ is the number of patients in \textit{each} class (positive and negative).
We use $k \in \{2,4,8,16,32,64,128\}$, and therefore have few-shot datasets \datasetfewshot{2}, \datasetfewshot{4}, \datasetfewshot{8}, \datasetfewshot{16}, \datasetfewshot{32}, \datasetfewshot{64}, and \datasetfewshot{128}.

\begin{table}[t]
\centering
\begin{adjustbox}{max width=\textwidth}
\begin{tabular}{@{}llp{2.5cm}p{2.3cm}p{2cm}p{3cm}@{}}
\toprule
\multirow{2}{1.4cm}{Dataset name} &\multirow{2}{2.6cm}{\# patient notes mean (min, max)} & \multicolumn{4}{c}{total patients (\#positive, \#negative)}                                                                                  \\
\addlinespace
                      && train & valid & test\_1 & test\_2 \\
\midrule
\datasetbalanced{}    & 32.2 (1, 284) & 1384 (692, 692)  & 518 (259, 259) & 170 (85, 85) & 1394 (697, 697)                 \\
\cmidrule{2-6}
$\mathcal{D}^{1:10}$  & 28.1 (1, 293) & ---                 & ---                 & ---                  & 7667 (697, 6970)                 \\
$\mathcal{D}^{1:100}$ & 29.2 (1, 572 & ---                 & ---                 & ---                  & 70396 (697, 69699)                 \\
$\mathcal{D}^{1:250}$ & 29.5 (1, 1403) & ---                 & ---                & ---                  & 174946 (697, 174249)                 \\
\cmidrule{2-6}
\datasetfewshot{2}    & 30.7 (1, 284) & 2 (1, 1)                      & 2 (1, 1)                & ---                  & ---                  \\
\datasetfewshot{4}    & 30.7 (1, 284) & 4 (2, 2)                      &  4 (2, 2)             & ---                 & ---                  \\
\datasetfewshot{8}    & 30.9 (1, 284) & 8 (4, 4)                      & 8 (4, 4)                 & ---                  & ---                \\
\datasetfewshot{16}   & 30.9 (1, 284) & 16 (8, 8)                     & 16 (8, 8)                 & ---                  & ---                \\
\datasetfewshot{32}   & 30.9 (1, 284) & 32 (16, 16)                   & 32 (16, 16)                & ---                  & ---                  \\
\datasetfewshot{64}   & 31.1 (1, 284) & 64 (32, 32)                   & 64 (32, 32)              & ---                  & ---                 \\
\datasetfewshot{128}  & 31.5 (1, 284) & 128 (64, 64)                  & 128 (64, 64)                 & ---                  & ---                 \\
\bottomrule
\end{tabular}
\end{adjustbox}
\caption{Dataset statistics: we show characteristics for our balanced dataset split \datasetbalanced{}, our imbalanced test sets $\mathcal{D}^{pos:neg}$, and our few-shot datasets \datasetfewshot{k}. For each dataset, we show the number of patients in the \textit{train}, \textit{valid}, \textit{test\_1} and \textit{test\_2} splits, as well as the mean, minimum and maximum number of free-text notes associated to each patient. Overall, there are a total of \textit{1,733} positive patients in our cohort (692+259+85+697).}\label{tab:data_statistics}
\end{table}

\subsection{Aggregating per-note into per-patient predictions}
\label{sec:note_to_patient_predictions}
In our cohort introduced in Section~\ref{sec:data}, patients have on average $28.1$--$32.2$ medical notes in their medical history (see Table~\ref{tab:data_statistics}), where each note has on average $26.2$--$34.2$ tokens.
Both PLMs we use in this work, \mbox{MedRoBERTa.nl} and RobBERT, process texts with a maximum length of $512$ tokens.
Nonetheless, we wish to build models that  \textit{predict risk of lung cancer for a patient} and thus use all medical notes in the patient's medical history as predictors.
In practice, we train our models to predict lung cancer \textit{independently from each note} available for a patient, and apply a simple but effective \textit{post-hoc aggregation} step to compute per-patient predictions from per-note predictions.

First, we construct each training instance $\{ X^P_n, Y^P\}_{n=1}^N$ by propagating the label $Y^P$ of patient $P$ to each of the $N$ notes available for that patient, i.e., $X^P_n$ is the $n$-th note belonging to patient $P$, and $Y^P$ is a binary variable indicating whether patient $P$ has lung cancer.
After the model is trained, we aggregate per-note into per-patient predictions following Equation~\ref{eq:note_to_patient_predictions} below.
\begin{equation}\label{eq:note_to_patient_predictions}
    P(Y^P=1 | \{X^P_n\}_{n=1}^N) = P_{min},
\end{equation}

\noindent
where $P_{min}$ is the lowest per-note probability predicted across all notes $\{X^P_n\}_{n=1}^N$.

\subsection{Soft-prompt tuning}
\label{sec:prompt_tuning}
Soft-prompt tuning is a method of model adaptation where a small number of new parameters are added to a model throughout that model's architecture, as opposed to standard model fine-tuning where a classification layer is typically appended at the ``end'' of the model
\cite{liu_gpt_2021,li_prefix-tuning_2021,liu-etal-2022-p}.
Concretely, we append \textit{continuous, trainable embeddings} to the input sequence which are optimised via backpropagation, while the original model parameters remain frozen and are not updated.
We refer the reader to~\cite{liu-etal-2022-p} for details.

\subsection{Models}
\label{sec:plms}

\subsubsection{Pretrained language models (PLMs)}
We use two Dutch-language PLMs based on the RoBERTa architecture \cite{liu_roberta_2019}:
\mbox{RobBERT}  \cite{delobelle-etal-2020-robbert}, which was created by retraining the RoBERTa tokenizer and model from scratch on a Dutch general language corpus, and \mbox{MedRoBERTa.nl} \cite{Verkijk_Vossen_2021}, which was created with a similar procedure, but using Dutch hospital notes instead of general language data. 

\subsubsection{Static word embedding models (WEMs)}
FastText~\cite{bojanowski2016enriching} is a static WEM where word representations encode \textit{subword information}.
After FastText has been trained, however, the meaning of a word will \textit{not} change when changing its context.
However, FastText shows strong results across text classification tasks, especially in tasks where noisy input texts are common, e.g. texts containing typos, mispellings, and acronyms~\cite{joulin2016bag}.
For that reason, we use FastText as one of the baselines in our experiments.

We follow two steps:
1) We pretrain FastText from scratch using all free-text notes available for all patients from all splits in \datasetbalanced{}.
2) We use the training notes in \datasetbalanced{} to fine-tune FastText to predict risk of lung cancer for that dataset.

\section{Results and Discussion}
\label{sec:results}

{
\setlength{\tabcolsep}{10pt} % Default value: 6pt
\renewcommand{\arraystretch}{1.5} % Default value: 1
\begin{table}[t]
\centering
\begin{adjustbox}{max width=0.8\textwidth}
\begin{tabular}{@{}lllllll@{}}
\toprule
\multirow{2}{*}{Model} & \multicolumn{2}{l}{AUROC ($\uparrow$)} & \multicolumn{2}{l}{AUPRC ($\uparrow$)} & \multicolumn{2}{l}{Brier score ($\downarrow$)} \\
                       & N           & P           & N           & P           & N              & P              \\
\midrule
%\multicolumn{7}{c}{\bf Standard fine-tuning} \\
%\midrule
FastText (FT)               & ---            & $90.8$        & ---           & $\underline{90.4}$        & ---              & $\bm{44.7}$ \\
RobBERT (FT)                & $81.8$         & $90.2$        & $65.7$        & $84.8$        & $18.2$           & $\underline{54.3}$            \\
MedRoBERTa.nl (FT)          & $82.4$         & $90.2$        & $68.6$        & $82.7$        & $17.6$           & $56.8$           \\
%\midrule
%\multicolumn{7}{c}{\bf Soft-prompt tuning} \\
%\midrule
\cmidrule{2-7}
RobBERT (ST)                & $\bm{84.5}$        & $\bm{94.3}$        & $\underline{72.3}$        & $\bm{91.4}$        & $\underline{16.7}$           & $55.9$           \\
MedRoBERTa.nl (ST)          & $\underline{84.3}$        & $\underline{92.7}$        & $\bm{75.5}$        & $86.5$        & $\bm{15.9}$           & $58.0$           \\
\bottomrule
\end{tabular}
\end{adjustbox}
\caption{%Static word embedding model \mbox{FastText} and pretrained language models \mbox{RobBERT} and \mbox{MedRoBERTa.nl} 
Results on \datasetbalanced{} \textit{test\_1} for models fine-tuned (FT) to predict lung cancer vs. using soft-prompt tuning (ST). We highlight best results in \textbf{bold} and second-best by \underline{underscoring}. \textbf{N} denotes per-note and \textbf{P} per-patient results. We aggregate per-note into per-patient predictions following Equation~\ref{eq:note_to_patient_predictions}.}\label{tab:balanced_dataset}
\end{table}
}

\subsection{Full data availability and balanced classes}
\label{sec:res:full_data}
In Table~\ref{tab:balanced_dataset}, we report results on our experiments where we use the \datasetbalanced{} dataset.
The \datasetbalanced{} dataset includes all the positive patients in our cohort and an equal number of negative patients sampled randomly, i.e., we build a version of the data set with balanced positive/negative classes (see Table~\ref{tab:data_statistics} for more details).
Here, we aim to investigate how the methods used in this work fare in this idealised scenario where class imbalance is not an issue.
Moreover, we compare models when directly \textit{fine-tuned} or when \textit{soft-prompt tuned} on the task of predicting whether a patient has lung cancer.
We use the \textit{train}, \textit{valid}, and \textit{test\_1} splits for model training, model selection, and testing, respectively.

\subsubsection{From per-note to per-patient predictions}
From Table~\ref{tab:balanced_dataset}, we first highlight that per-patient predictions---which aggregates per-note predictions for a patient according to Equation~\ref{eq:note_to_patient_predictions}---consistently improve on per-note predictions, which suggests that the offline aggregation method we use can suppress possibly noisy per-note predictions well.
We also experimented with aggregating per-note into per-patient predictions as 1) the mean, 2) the maximum, and 3) Clinical BERT~\cite{clinicalbert}'s aggregation over all note probabilities, however using Equation~\ref{eq:note_to_patient_predictions} performs best (we do not report those other experiments due to space constraints).

\subsubsection{Discussion}
First of all, we note that FastText performs comparably to or better than our \textit{fine-tuned} PLMs according to all metrics;
however, when compared to our \textit{soft prompt-tuned} PLMs, FastText is slightly worse (AUROC), comparable (AUPRC) or much better in terms of calibration (Brier score).
Moreover, soft-prompt tuning tends to consistently outperform standard model fine-tuning, according to AUROC (by $\sim 2$--$4\%$) and AUPRC (by $\sim 4$--$6\%$), but with no clear trend according to the Brier score.

Finally, per-note Brier scores are considerably better than per-patient Brier scores.
This suggests that the aggregation method we use (see Equation~\ref{eq:note_to_patient_predictions}) consistently improves AUROC and AUPRC at the expense of model calibration.

\subsection{Balanced vs. imbalanced classes}
\label{sec:res:imbalanced}

In Table~\ref{tab:imbalanced}, we show results for our FastText baseline and the best performing RobBERT model trained on \datasetbalanced{} (\textit{train}) and selected according to AUROC scores on \datasetbalanced{} (\textit{valid}).
We provide results on the \textit{test\_2} splits of \datasetbalanced{}, $\mathcal{D}^{1:10}$, $\mathcal{D}^{1:100}$, and $\mathcal{D}^{1:250}$.

As expected, we note that both AUROC and AUPRC tend to decrease as we increase the number of negative patients, with more pronounced effects on AUPRC (i.e., when test sets have roughly 1:250 patient-to-positive ratios, AUPRC scores are below $1\%$).
However, we also note an opposite trend: \textit{Brier scores improve as test sets become more imbalanced}.
This makes sense when taking our per-patient aggregation method into account: since we use the lowest note probability, the prediction for true negatives will likely be much closer to 0 than the prediction for true positives will be to 1. 

{
\begin{table}[t]
\centering
\begin{adjustbox}{max width=\textwidth}
\begin{tabular}{@{}l@{\kern2em}l@{\kern1em}l@{\kern1em}l@{\kern1em}l@{\kern2em}l@{\kern1em}l@{\kern1em}l@{\kern1em}l@{\kern2em}l@{\kern1em}l@{\kern1em}l@{\kern1em}l@{}}
\toprule
 & \multicolumn{4}{c}{AUROC ($\uparrow$)} & \multicolumn{4}{c}{AUPRC ($\uparrow$)} & \multicolumn{4}{c}{Brier score ($\downarrow$)} \\
\midrule
Test set $\mathcal{D}^?$     & $^{bal}$ & $^{1:10}$ & $^{1:100}$ & $^{1:250}$ 
      & $^{bal}$ & $^{1:10}$ & $^{1:100}$ & $^{1:250}$ 
      & $^{bal}$ & $^{1:10}$ & $^{1:100}$ & $^{1:250}$ \\
\midrule
Random & 50.0 & 50.0 & 50.0 & 50.0 & 50.0 & 10.0 & 1.0 & 0.4 & --- & --- & --- & --- \\
\midrule
\multicolumn{13}{c}{Per-note predictions} \\
\midrule
RB (FT)  & 76.8 & 78.3 & 61.3 & 54.4 &
                63.9 & 15.9 & 1.2  & $\bm{0.4}$  &
                19.9 & 18.7 & 28.3 & 32.4 \\
RB (ST)  & $\bm{80.1}$ & $\bm{81.4}$ & $\bm{64.5}$ & $\bm{57.8}$ &
                $\bm{70.6}$ & $\bm{22.0}$ & $\bm{1.5}$ & $\bm{0.4}$ &
                $\bm{18.3}$ & $\bm{16.8}$ & $\bm{27.4}$ & $\bm{32.3}$ \\
\midrule
\multicolumn{13}{c}{Per-patient predictions}\\
\midrule
FaT (FT) & 78.7 & 81.0 & 60.1 & 56.1 &
                77.2 & 32.3 &  1.4 &  0.5 &
                $\bm{44.8}$ &  $\bm{8.2}$ &  $\bm{1.2}$ &  $\bm{0.7}$ \\
RB (FT)  & 88.2 & 90.1 & 70.3 & 65.4 &
                81.9 & 40.0 & 1.8   & 0.6   &
                54.0 & 67.7 & 24.3  & 18.0 \\
RB (ST)  & $\bm{89.7}$ & $\bm{91.1}$  & $\bm{71.2}$ & $\bm{67.1}$  &
                $\bm{86.2}$ & $\bm{47.0}$  & $\bm{1.9}$  & $\bm{0.7}$ &
                56.3 & 71.0  & 25.1 & 17.8 \\
\bottomrule
\end{tabular}
\end{adjustbox}
\caption{Results on imbalanced vs. balanced test datasets with the best performing fine-tuning (FT) and soft prompt-tuning (ST) model (RobBERT) trained on \datasetbalanced{} \textit{train} split.
\textbf{N} denotes per-note and \textbf{P} per-patient results. {\bf RB}: RobBERT. {\bf FaT}: Fast-text. We aggregate per-note into per-patient results with Equation~\ref{eq:note_to_patient_predictions}.}\label{tab:imbalanced_dataset}
\label{tab:imbalanced}
\end{table}
}

\subsection{Few-shot learning}
\label{sec:res:few_shot}
Few-shot learning results show that 
when models are trained on 32 or less patients, PLMs tend to clearly outperform FastText, but when models are trained on 128 patients we already note that FastText outperforms all PLMs (according to both AUROC and AUPRC).
This happens regardless of the PLM being fine-tuned or soft-prompt tuned, which suggests that soft-prompt tuning is still not clearly more resilient to `noisy' inputs than fine-tuned models, at least according to our experiments.

Moreover, we note that fine-tuned and soft-prompt tuned models seem to perform comparably in our few-shot experiments.
Oscillations in both AUROC and AUPRC from smaller to higher $k$'s suggest that there is some degree of randomness in model performance, which is an undesirable feature which we would like to address in future work.

\begin{figure*}[t!]
    \centering
    \begin{subfigure}[t]{0.47\textwidth}
        \centering
        %\includesvg[height=6.5cm]{imgs_submission/auroc_patient_val2.svg}
        \includegraphics[height=6.5cm]{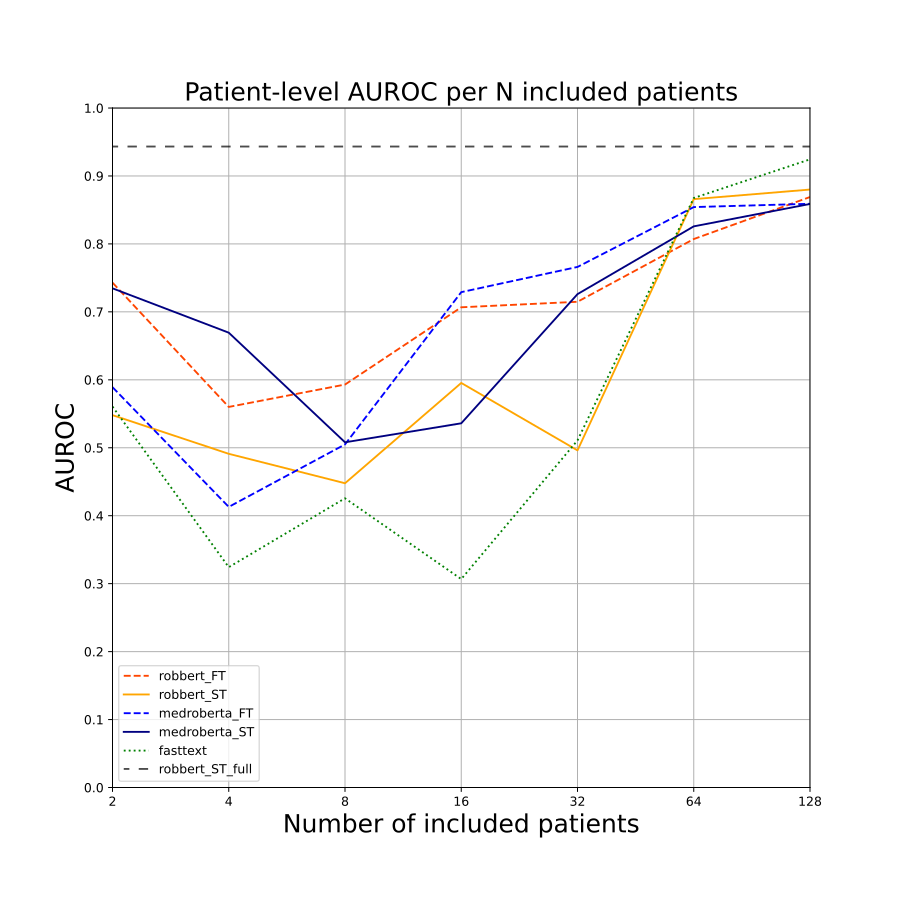}
        \caption{Number of training patients vs. \mbox{AUROC} scores on \datasetbalanced{} \textit{test\_1}.}
    \end{subfigure}\hfill%
    ~
    \begin{subfigure}[t]{0.47\textwidth}
        \centering
        %\includesvg[height=6.5cm]{imgs_submission/auprc_patient_val2.svg}
        \includegraphics[height=6.5cm]{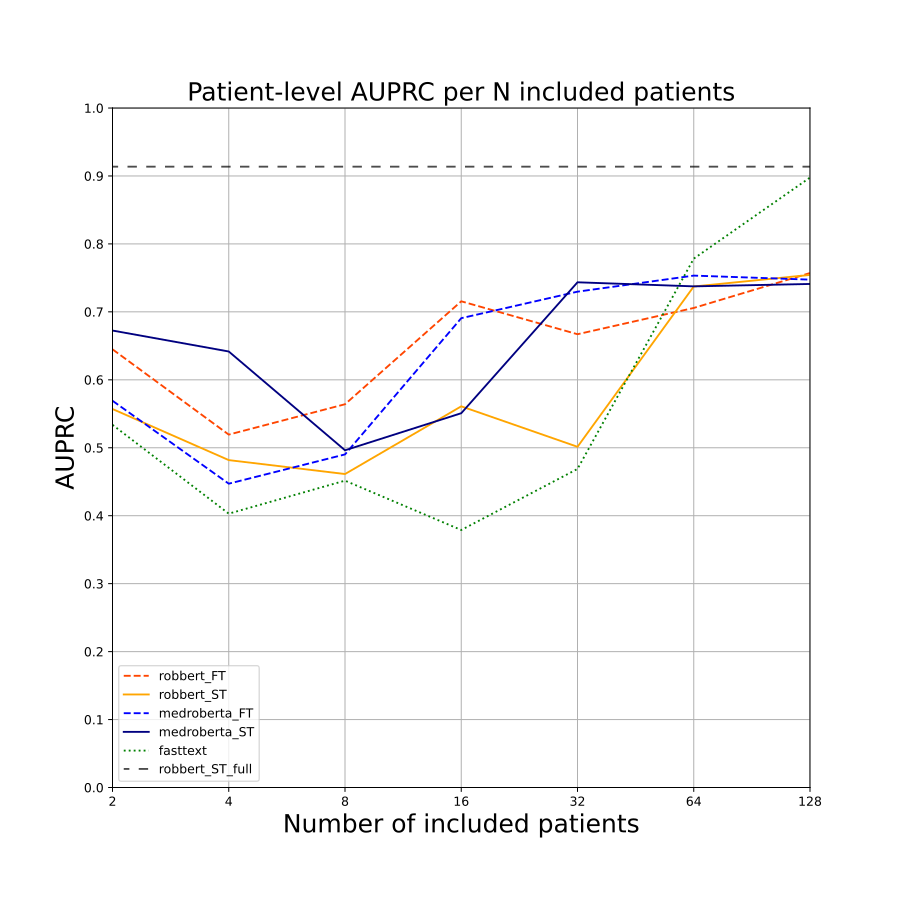}
        \caption{Number of training patients vs. AUPRC scores on \datasetbalanced{} \textit{test\_1}.}
    \end{subfigure}
    \caption{Results
    %on \datasetbalanced{} \textit{test\_1} 
    for models trained on \datasetfewshot{k} (\textit{train}), $k \in \{2, 4, 8, 16, 32, 64, 128 \}$. {\bf ST}: soft-prompt tuning. {\bf FT}: fine-tuning. RobBERT 'full' is the best RobBERT model trained and tested on \datasetbalanced{} \textit{train} and \textit{test\_1} splits, respectively.}
\end{figure*}
\section{Related Work}
\label{sec:related}

% Using NLP for medical data
Recent years have seen the release of many English-language PLMs (pre)trained on clinical data~\cite{golmaei_deepnote-gnn_2021,rasmy_med-bert_2021,li_hi-behrt_2021,kim_deep-learning-based_2021}.
Dutch-language PLMs are much more rare, and even more so in the medical domain: we are only aware of Verkijk and Vossen's PLM~\cite{Verkijk_Vossen_2021}, which is pretrained on hospital patient notes.
% Cancer prediction
Whereas a variety of models predict cancer incidence or survival based on hospital notes \cite{kim_deep-learning-based_2021,aken_clinical_2021}, we only know of Torec et al.~\cite{torec-et-al2021w2vlungcancer}'s model that use \textit{free-text GP notes} as predictors.
% (fewshot) Soft prompt tuning
In recent years, a large variety of prompt tuning methods have become available.
Prompt tuning approaches use a fraction of the data and compute compared to models that fine-tune all parameters in a PLM, which is one of the main reasons we investigate prompt tuning in our work~\cite{li_prefix-tuning_2021,liu_gpt_2021}.
\section{Conclusions and Future Work}
\label{sec:conclusions}
In this work, we investigate the application of soft-prompt tuning to perform the early prediction of lung cancer using free-text patient medical notes
of Dutch primary care physicians.
We find that using contextualised pretrained language models (RobBERT and \mbox{MedRoBERTa.nl}) outperforms strong static word embedding models (FastText) according to AUROC and AUPRC, however FastText shows much better calibration than both PLMs.
Results we obtained in few-shot experiments are not so clear cut, and here the difference between PLMs and WEMs is less clear.
Soft prompt tuning consistently outperforms standard model fine-tuning with PLMs, which we find promising and believe warrants further research.
When testing our models on datasets with increased class imbalance, performance deteriorates as expected; nonetheless, the best performing PLM still achieves a reasonable $67.1$ AUROC when tested on a $1$:$250$ (\textit{positive:negative}) ratio---virtually the same ratio as that of the population in our dataset---, though the corresponding $0.7$ AUPRC shows modest improvements upon the random baseline ($0.4$ AUPRC).

As future work, we plan to investigate soft prompt tuning techniques with PLMs and how to best combine these models with static word embedding models to try to obtain both high discrimination \textit{and} calibration.
We will also investigate how to stabilise predictions in few-shot experiments as we increase the number of training examples, possibly training models with different patient populations (but the same number of patients) as a form to estimate uncertainty.

\bibliographystyle{splncs04}
\bibliography{refs}

\end{document}